\PassOptionsToPackage{expansion=false}{microtype}
\documentclass[sigconf,nonacm]{acmart}

\usepackage{fontspec}
\usepackage{booktabs}
\usepackage{graphicx}
\usepackage{multirow}
\usepackage{float}
\usepackage{tikz}
\usetikzlibrary{arrows.meta,positioning,fit}

\newcounter{algnum}
\newenvironment{algblock}[1]{%
  \par\addvspace{\medskipamount}\refstepcounter{algnum}%
  \noindent\rule{\columnwidth}{0.8pt}\par\nobreak\vskip2pt%
  \noindent\textbf{Algorithm \thealgnum}\ #1\par\nobreak\vskip2pt%
  \noindent\rule{\columnwidth}{0.4pt}\par\nobreak\vskip3pt%
  \setlength{\parskip}{0pt}%
}{%
  \par\vskip3pt\noindent\rule{\columnwidth}{0.8pt}\par\addvspace{\medskipamount}%
}

\setcopyright{none}
\renewcommand\footnotetextcopyrightpermission[1]{}
\begin{document}

\title{LV-ROVER-MLT: Low-Resource Maltese OCR by Synthetic Fine-Tuning and Multi-Stream Arbitration}

\author{Adam Darmanin}
\affiliation{%
  \institution{Independent Researcher}
  \country{}
}
\email{adamdarmanin@hecatusresearch.com}

\renewcommand{\shortauthors}{Darmanin}

\begin{abstract}
Maltese has substantial text corpora and pretrained language models, but
paragraph-scale OCR training data remains scarce; NOMOCRAT provides 57
verified annotated pages. LV-ROVER-MLT combines synthetic fine-tuning of
Tesseract~5 with five complementary recognition streams and
lexicon-gated word-level arbitration adapted to Maltese diacritics and
hyphenation. In the DocEng~2026 Maltese OCR competition, the system
placed first with held-out CER 0.0074; the next-ranked submission scored
0.0161 and NOMOCRAT scored 0.0163. The same approach produced a
significant improvement over stock Tesseract on Luxembourgish, while
the Hungarian result was inconclusive. A 36,803-pair Maltese OCR corpus
constructed from EUR-Lex and Wikipedia provides an additional
paragraph-level resource. Code, model weights, and corpus data are public.
\end{abstract}

\keywords{OCR, low-resource, Maltese, synthetic data, character error rate}

\maketitle

\section{Introduction}

Maltese is a Semitic language written in Latin script, spoken by
approximately half a million native speakers. Decades of archival text in
Maltese remain unsearchable because they have not been digitised at scale,
and OCR is the prerequisite step.

The language's script is a difficult OCR target: its 30-letter alphabet extends the
standard Latin alphabet with \textit{Ċ ċ}, \textit{Ġ ġ}, \textit{Ħ ħ},
\textit{Ż ż}, and the digraphs \textit{Għ għ} and \textit{Ie ie}; fonts
silently substitute the base letter for the diacritic form; and the definite
article attaches to the following noun via a structural hyphen
(\texttt{il-kelb}, \texttt{id-dar}, \texttt{fis-seħħ}) that shares the same
glyph, \texttt{-}, with soft line-break hyphens. The only labelled real
Maltese PDF data we are aware of comes from the NOMOCRAT annotation
project~\cite{tanti2023nomocrat}: 57 PDF pages, line- and
paragraph-transcribed per the project's own description. That is far
below what training a paragraph-level recogniser needs.

Maltese OCR presents several language-specific failure modes. Fonts and
tokenisation pipelines can silently replace \textit{ċ}, \textit{ġ},
\textit{ħ}, or \textit{ż} with their ASCII bases, corrupting labels
before recognition is evaluated. We therefore track these four
diacritic pairs throughout data generation and inference. Hyphenation
introduces a different ambiguity: structural forms such as
\texttt{il-kelb} and \texttt{fis-seħħ} use the same glyph as
line-breaking hyphens, so resolution requires Maltese-aware
joining rather than unconditional hyphen removal
\cite{borg2017morphology,mlrs2025maltipkg}. The benchmark also uses
curly quotation marks and em-dashes where raw Tesseract output commonly
uses ASCII forms. We therefore distinguish recognition improvements
from agreement with the benchmark's output convention.

The DocEng~2026 evaluation environment imposes no network access, a
fixed disk budget, a runtime ceiling, Python~3.9, and an 11GB
RTX~2080~Ti. These constraints favour compact
recognisers and deterministic inference; LV-ROVER-MLT therefore uses
Tesseract streams rather than a large neural decoder such as
TrOCR \cite{li2023trocr}.

The contributions are:
\begin{itemize}
\item A synthetic Maltese OCR training procedure that preserves
language-specific diacritics and reproduces the resolution, compression,
and hyphenation characteristics of the competition documents.

\item An explicit separation between recognition error and
benchmark-specific transcription normalization, allowing improvements in
visual recognition to be distinguished from quote and dash convention
matching.

\item A lexicon-gated multi-stream arbitration method that preserves
open-vocabulary recognition while recovering Maltese diacritics from
complementary Tesseract hypotheses.

\item A public 36{,}803-pair paragraph-level Maltese OCR corpus derived
from EUR-Lex and Wikipedia with per-sample provenance and licensing.
\end{itemize}

\section{Related Work}
\label{sec:related}

The Maltese Language Resource Server maintains Korpus Malti with POS,
lemma, and morphological annotation~\cite{micallef2022pretraining,
borg2017morphology}, but a shortage of labelled OCR image-text data at
paragraph scale remains. Prior work on Maltese OCR is limited to
NOMOCRAT~\cite{tanti2023nomocrat}, which surfaced the hyphen-joining and
label convention challenges this system addresses.

Consensus-sequence voting over aligned OCR outputs was established for text recognition in the
1990s~\cite{lopresti1997consensus}, and later work aligned and combined
multiple full OCR systems directly~\cite{lund2009multiple}. Modern OCR
engines fold voting inside the recogniser: Calamari votes across
independently trained networks by per-character
confidence~\cite{wick2020calamari}, the engine-native form of the idea this
system applies across Tesseract configurations. ROVER's word-level voting
across independent recogniser outputs~\cite{fiscus1997rover} is the general
mechanism this system adapts. LV-ROVER \cite{stuner2020lvrover} is the closest methodological
predecessor. It combines complementary LSTM recognisers using
lexicon-verified voting and was evaluated on RIMES with recognisers
drawn from multiple architectures, checkpoints, and initialisations.
LV-ROVER-MLT retains lexicon-verified arbitration but obtains diversity
from language-chain, model-source, and image-scale variation, reflecting
the smaller training resources and fixed execution budget of the
Maltese setting.

\subsection{Adaptation of LV-ROVER}
\label{sec:rover-adapt}

LV-ROVER-MLT retains word-level alignment and hypothesis arbitration
from ROVER and LV-ROVER, but modifies how candidate outputs can alter
the recognised paragraph.

One anchor stream fixes the output structure and the other streams
propose only substitutions at anchor-aligned positions; they cannot insert
or delete to restructure the anchor, unlike ROVER's symmetric confusion
network. The lexicon is a membership eligibility gate not a frequency table or a closed vocabulary, so an out-of-vocabulary
anchor word survives unless a candidate stream offers an eligible
replacement.

A diacritic-restoration gate extends LV-ROVER's lexicon
verification to the Maltese setting, promoting a diacritic-richer
in-lexicon candidate even over an already-valid anchor. Arbitration is
plurality with no quorum: the most-voted eligible proposal wins, a single
eligible distinct hypothesis can carry a position, and ties break by
stream order, not majority.

\section{Method}

\subsection{System Overview}

Synthetic Maltese text is rendered into line images used to fine-tune
the Tesseract LSTM. At inference time, five deterministic Tesseract
configurations recognise the same paragraph image and their outputs are
combined by anchor-preserving, lexicon-gated arbitration.

The five-stream configurations obtain diversity without retraining: a stream fails
on different characters than its siblings only if its inputs differ.
Diversity here comes from language chain (Maltese alone, Maltese plus
Italian, Maltese plus Italian and French), training data (fine-tuned versus
stock recogniser), and image scale (native versus 2x-upscaled crop). Each
axis changes the recogniser's error profile, creating complementary
hypotheses that can be recovered through arbitration.

\begin{figure*}[t]
  \centering
  \includegraphics[width=0.78\textwidth]{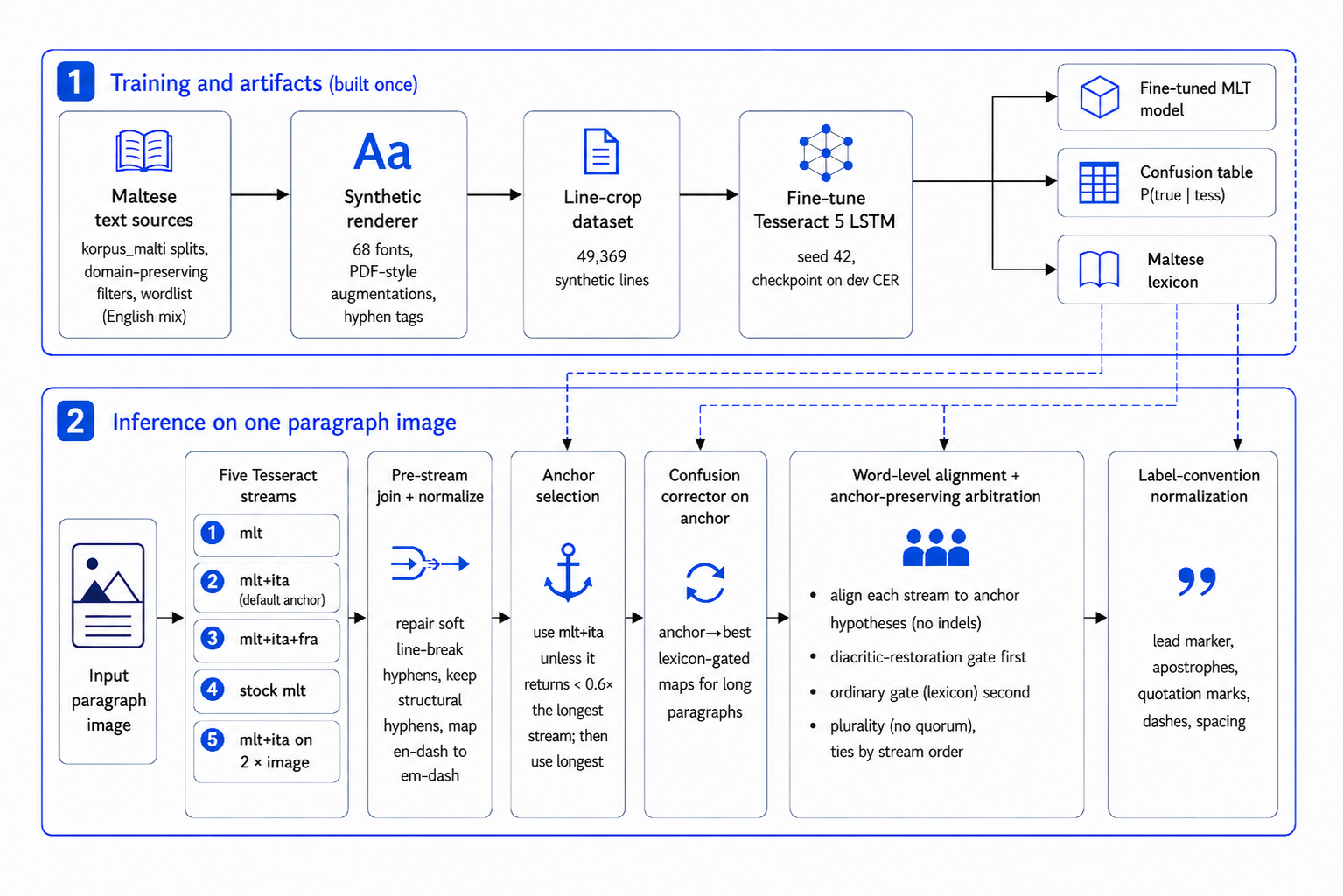}
  \caption{LV-ROVER-MLT training (top) and five-stream inference with
  anchor-preserving arbitration (bottom).}
  \label{fig:pipeline}
\end{figure*}

Figure~\ref{fig:pipeline} summarizes the training and inference
pipeline. Synthetic text produces the fine-tuned model, character
confusion statistics, and Maltese lexicon. The
\texttt{mlt+ita} stream provides the default anchor, with a
longest-output fallback for incomplete recognition. Candidate streams
are aligned to anchor words, filtered by the lexicon and
diacritic-restoration rules, and combined without changing the anchor's
paragraph structure. Quote and dash normalization is applied after
arbitration.

\textit{Ghadha} is the anchor. Two hypotheses produce \textit{Għadha},
lexicon-valid and differing only by diacritic restoration. A third
matches the anchor and casts no vote. \textit{Għada} is rejected for
shortening the alphabetic form. The eligible plurality selects
\textit{Għadha}.

\subsection{Synthetic Data Pipeline}

Because labelled Maltese paragraph OCR data is scarce, model
fine-tuning relies on synthetically rendered Maltese text.

\subsubsection{Text Source}
\label{sec:textsource}

Training text comes from korpus\_malti
v4.2~\cite{mlrs_korpus_malti_card,micallef2022pretraining}, which
contains approximately 470M tokens across 19 domain-specific
configurations. Eleven configurations retain document-level sentence
order and sufficient Maltese diacritic coverage:
\texttt{parliament}, \texttt{wiki}, \texttt{government\_gazzette},
\texttt{law\_mt}, \texttt{nonfiction}, \texttt{theses},
\texttt{legal}, \texttt{speeches}, \texttt{blogs},
\texttt{umlib\_oar}, and \texttt{web\_general}. The global shuffled
configuration is excluded because it breaks paragraph coherence, while
\texttt{press\_mt} is excluded because its diacritic density is
approximately 0.01 compared with 0.12--0.20 for the retained sources.
Wikipedia Maltese and the Maltese Universal Dependencies treebank
provide fallback text, and English code-switch material contributes
12\% of the synthetic text.

\subsubsection{Synthetic Image Renderer}
\label{sec:renderer-dpi}

The renderer is a SynthTIGER compatible wrapper~\cite{yim2021synthtiger}
using Pillow\footnote{\url{https://python-pillow.org/}} for paragraph
layout. Each rendered sample emits an image, a
label string, per-line label parts with a soft-hyphen marker (U+00AD, the
invisible soft hyphen that renders only at an actual line break) at every
line-break position, and metadata recording per-line bounding boxes, font,
and hyphen kind.

Resolution is calibrated to the benchmark distribution. Synthetic-to-real
scale mismatch is a known source of recognition error
\cite{gupta2016synthtext,baek2019wrong}; the competition crops are
approximately 150--200~DPI, whereas unscaled 300~DPI rendering produces
characters roughly twice the required size. The renderer therefore
applies a half-resolution Lanczos rescale and JPEG re-encoding at
quality~72 to approximate the benchmark crops.

PDF-realistic augmentations are applied as a fixed chain: rotation, blur,
brightness/contrast jitter, optional ink bleed and column-edge crop, mild
elastic distortion, salt-and-pepper noise, JPEG re-encoding, block noise,
page-edge shadow, and subpixel blur. Scene-text operations (perspective
warp, motion blur, glare) are disabled; they do not occur in PDF crops.

The renderer tags every hyphen it draws as soft (line-break split),
structural (clitic-article surfaces such as \textit{il-} and
\textit{fis-}), or compound (word-internal), which lets us measure joiner
accuracy on soft hyphens alone without contamination from structural
hyphens, which must be preserved rather than removed.

\subsubsection{Font Catalogue}

A font can silently substitute the base Latin letter for a Maltese diacritic
at render time. We curated 68 faces (62
printed, 6 handwriting) under permissive licences (SIL Open Font License,
GUST, Apache~2.0, DejaVu). Each candidate is checked for character-map
presence of the canary glyphs \textit{Ċ ċ Ġ ġ Ħ ħ Ż ż à ì ò ù} before
entering the renderer pool; five candidates failed this check, missing one
or more of \textit{Ċ ċ Ġ ġ Ħ ħ}. This check catches a glyph that is absent
outright; it does not verify that a present glyph renders visually distinct
from its ASCII base, the specific risk noted in prior low-resource OCR
work. The validation therefore confirms glyph availability but not visual
distinctness from the corresponding ASCII base character.

Joiner behaviour is evaluated on independently generated synthetic
paragraph sets, including a fully augmented held-out set with explicit
hyphen tagging.

\subsection{Recogniser and Baseline Anchor}
\label{sec:model}

The recogniser is Tesseract~5~\cite{smith2007tesseract,
smith2009multilingual}, fine-tuned on a separate corpus of 49{,}369
synthetic line crops generated from five rendering shards. With the
rule-based joiner, it reaches CER~0.01605 on the development set by itself.
This is our fine-tuned anchor, distinct from the organizers' fine-tuned
Tesseract baseline (CER~0.0234).

Canary diacritics and the two label-bearing dashes
round-trip through a 117-character training inventory. Structural
clitic-article hyphens (\textit{il-}, \textit{is-}, \textit{id-},
\textit{fis-}, and others) are never removed by the joiner's line-break
repair; the image-only en-dash (U+2013) is normalised to an em-dash
(U+2014). Tesseract is deterministic; inference runs on CPU.

Fine-tuning starts from Tesseract's stock Maltese model, warm-started onto
a merged character set (stock plus the 117-character training inventory).
Training runs 2,000 iterations at seed~42 over five shards with a 70/30
augmented-to-clean mix and checkpoints every 100 iterations. Iteration~1400
is selected by minimum CER on the 422-paragraph development set, scoring
0.016053; performance is broadly flat from iteration~400 onward. Because
no independent validation partition is used for checkpoint selection, this
development-set estimate is not an independent validation result.

\subsection{Joining and Lexicon-Gated Arbitration}
\label{sec:arbitration}

\subsubsection{Joiner}

The joiner resolves soft-versus-structural hyphens at decode
time~\cite{mlrs2025maltipkg} using rule-based hyphenated-word repair on
each stream's per-line output. Its result is then
normalised: the soft-hyphen marker (U+00AD) is stripped and every en-dash
(U+2013) is converted to an em-dash (U+2014); gold always uses the em-dash.
A worked example: the image shows \textit{0~--~Għadha mhux fis-} /
\textit{seħħ}, and gold reads \textit{0~---~Għadha mhux fis-seħħ}. The
image-rendered en-dash maps to an em-dash in the label; the structural
\textit{fis-} hyphen is preserved across the line wrap; the soft hyphen is
removed and the word rejoined. The two label-bearing dashes, the ASCII
hyphen and the em-dash, are never normalised against each other.

The joiner achieves 99.51\% round-trip accuracy on synthetic
multi-line paragraphs and correctly resolves all evaluated soft-hyphen
cases; the remaining errors occur primarily at numbered-bullet line
starts.

\subsubsection{Anchor Selection and Confusion Correction}

The default anchor is the mlt+ita fine-tuned stream. If it returns fewer
than 0.6 times the longest stream's length, the longest stream becomes the
anchor instead. For paragraphs of 100 characters or more, a confusion
corrector then takes each anchor word not in the lexicon and tries
single-character swaps ranked by a character confusion table,
$P(\mathrm{true}\mid\mathrm{tess})$, estimated from synthetic
(rendered-character, recognised-character) pairs. It accepts a swap only
when the result is itself a lexicon word, is within edit distance~2, and
does not shorten the word's letter or diacritic count. The lexicon is fixed
throughout: the corrector maps a misread word onto an existing entry and
never adds to the lexicon.

\subsubsection{Lexicon-Gated Plurality Arbitration}

The remaining distinct non-anchor hypotheses are aligned word-by-word
against the anchor by edit distance and combined with an arbitration rule
adapted from
LV-ROVER~\cite{stuner2020lvrover}, itself an extension of
ROVER~\cite{fiscus1997rover}. ROVER aligns multiple recogniser outputs into
a confusion network and votes at corresponding positions, while LV-ROVER
adds lexicon verification. LV-ROVER-MLT retains this principle but uses two
Maltese-specific eligibility rules: the lexicon gate and the
diacritic-restoration gate.

The lexicon acts as an eligibility constraint rather than a closed
recognition vocabulary.
For each anchor word, every aligned candidate is checked against two gates
in order.

The diacritic-restoration gate fires first: if
a candidate's diacritic-stripped form matches the anchor's, the candidate
carries strictly more canary diacritics, the candidate is itself
lexicon-valid, and the two are within edit distance~2, that candidate is
accepted as a diacritic-restoring proposal even when the anchor is already
lexicon-valid. Only if this gate does not fire does the ordinary swap gate
apply: an anchor word already in the lexicon is otherwise kept, and an
out-of-lexicon anchor is replaced by a candidate only when that candidate
is itself in the lexicon, is within edit distance~2, does not shorten the
anchor's alphabetic length or drop any of the four canary characters it
carries, and does not shrink the anchor's non-ASCII letter count. Each distinct
surviving candidate that passes a gate casts one vote for its proposed
word; the proposal with the most votes replaces the anchor, with ties
broken by
stream order.

A single eligible proposal may replace the anchor when no competing
proposal exists, so arbitration requires plurality rather than a
majority quorum. The restoration rule promotes an in-lexicon form that
differs from the anchor only by adding Maltese diacritics
(e.g.\ \textit{zwieg}$\rightarrow$\textit{żwieġ}), even when the
anchor itself is lexicon-valid. The ordinary swap rule separately
prevents candidates from removing existing non-ASCII characters, such
as \textit{ħ}$\rightarrow$\textit{h}. Together, these constraints
extend LV-ROVER's lexicon verification to Maltese diacritic recovery.

Identical complete non-anchor hypotheses are collapsed before alignment,
so arbitration operates over distinct candidate hypotheses.

\subsection{Label-Convention Post-Processing}
\label{sec:postproc-method}

After arbitration, a three-pass chain converts Tesseract's output to the
gold label convention: a leading clause-marker rule (digit, dash, and space
around an em-dash at paragraph start), an apostrophe-curling rule, and a
positional double-quote rule that curls opening and closing double quotes
in a single pass. The diacritic-restoration step is not part of this
chain; it runs inside arbitration (Section~\ref{sec:arbitration}), before
these passes.

\section{Development-Set Evaluation}
\label{sec:results}

Results in this section are measured on the competition's 422-paragraph
development set under the DocEng evaluation constraints and are distinct
from the held-out competition results.

\subsection{Pre-Convention versus Label-Convention Gains}

Table~\ref{tab:five-stream-evals} shows the effect of arbitration before
benchmark-specific normalization. Five-stream arbitration improves on
the fine-tuned anchor, and adding the diacritic-restoration rule produces
a further reduction. For the restoration configuration, the
anchor-to-ensemble improvement is 0.00386 under paired resampling
(95\% CI 0.00266--0.00517; permutation $p<0.0001$).

\begin{table}[H]
\caption{CER before label-convention.}
\label{tab:five-stream-evals}
\resizebox{\columnwidth}{!}{%
\begin{tabular}{@{}lrrl@{}}
\toprule
Configuration & Anchor & CER & Delta \\
\midrule
Five-stream arbitration & 0.01605 & 0.01317 & 0.00288 \\
Five-stream + restoration & 0.01605 & 0.01220 & 0.00386 \\
\bottomrule
\end{tabular}}
\end{table}

The complete pipeline, including the diacritic-restoration rule and
benchmark-specific quote and dash normalization, reaches development-set
CER~0.00700. Because these normalization rules alter agreement with the
reference transcription rather than visual character recognition, the
pre-convention and full-pipeline results are reported separately.

Table~\ref{tab:versions} reports CER for the evaluated component
configurations; the differences are descriptive rather than independent
causal estimates.

\begin{table}[H]
\caption{Component results.}
\label{tab:versions}
\begin{tabular}{@{}llr@{}}
\toprule
Stage & Type & CER \\
\midrule
3-stream arbitration & recognition & 0.01441 \\
5-stream arbitration$^\dagger$ & recognition & 0.01317 \\
Lead-marker norm. & convention & 0.01294 \\
Apostrophe norm. & convention & 0.00810 \\
Opening-quote rule & convention & 0.00776 \\
Diacritic-restore gate & recognition & 0.00741 \\
Closing-quote norm. & convention & 0.00700 \\
\bottomrule
\end{tabular}
{\par\noindent\scriptsize $^\dagger$Statistical testing applies to the
anchor-to-five-stream result with diacritic restoration; intermediate
component differences are reported descriptively.}
\end{table}

\subsection{Ensemble Construction}
\label{sec:arms}

The recogniser streams are useful only when their errors are
complementary. Mean pairwise word disagreement is 4.0\%, while an
oracle that selects the best complete stream for each paragraph reaches
CER~0.01118 compared with 0.01432 for the best fixed stream. The
five-stream configuration with diacritic restoration reaches 0.01220,
close to this paragraph-level oracle despite operating at word level.
This indicates that the recognisers expose useful alternative
hypotheses for arbitration.

\subsection{CER Computation and Stratification}
\label{sec:cer-recipe}

CER is computed with the \texttt{jiwer} library~\cite{vaessen2025jiwer} on
NFC-normalised reference and hypothesis text, aggregated as a sum of edit
distances over a sum of reference lengths across paragraphs $p$:
\[
\mathrm{CER} = \frac{\sum_p \mathrm{edit}(ref_p, hyp_p)}{\sum_p \mathrm{len}(ref_p)}.
\]
This matches \texttt{jiwer}'s default aggregation and the organizers'
scoring script.

The regression criterion is evaluated across buckets defined by
paragraph length, language tag, clitic-article presence, line-break
hyphenation, and em-dash presence; buckets containing fewer than
20 paragraphs are excluded.

\subsection{Statistical Method}
\label{sec:audit}

Statistical analysis focuses on the anchor-to-five-stream improvement.
Uncertainty is estimated by paired percentile bootstrap over
per-paragraph edit distances using 1,000 resamples
\cite{efron1993bootstrap,koehn2004sigtest}. Significance is also tested
with 10,000 paired label-swap permutations using the add-one correction
$p=(n_{\ge}+1)/(N+1)$. Held-out synthetic evaluations provide a
stability check under the renderer distribution. Per-character
bootstrap tests with BH-FDR correction at $\alpha=0.05$ provide an
additional development safeguard for the Maltese diacritic set
\cite{benjamini1995controlling}.

Evidence of improvement requires a paired-resampling interval above zero
with no regression greater than 0.005 absolute CER in a sufficiently
populated evaluation bucket. All resampling is seeded (seed 42).

\subsection{Post-Processing Rule Audit}
\label{sec:rule-audit}

The normalization rules were evaluated by leave-one-out ablation on the
development set and on two held-out synthetic sets. The ablation tests
whether a development-set improvement is accompanied by an opposing
degradation on either synthetic evaluation set.

\begin{table}[H]
\caption{Leave-one-out ablation.}
\label{tab:audit}
\resizebox{\columnwidth}{!}{%
\begin{tabular}{@{}lrrrl@{}}
\toprule
Rule & Dev set & Synth A & Synth B \\
\midrule
Lead-marker norm. & +0.00022 & +0.00000 & +0.00000 \\
Apostrophe norm. & +0.00496 & \emph{n/a} & \emph{n/a} \\
Opening-quote pos. & +0.00025 & +0.00000 & +0.00000 \\
Diacritic-restore gate & +0.00037 & +0.00074 & +0.00063 \\
Closing-quote norm. & +0.00000 & \emph{n/a} & \emph{n/a} \\
\bottomrule
\end{tabular}}
{\scriptsize\textit{n/a}: synthetic and benchmark quotation conventions differ.}
\end{table}

The quotation-normalization rows are not comparable under the synthetic
label convention. The remaining evaluated rules show no opposing
degradation on either synthetic set.

\section{Additional Evidence}
\label{sec:postsub}

\subsection{Cross-Language Portability}
\label{sec:crosslang}

We evaluated portability of the same fine-tuning-plus-arbitration protocol
to two other languages, $n=200$ paragraphs per language: Hungarian
(HuCCPDF~\cite{li2026huccpdf}) and Luxembourgish (19th-century newspaper
print). Neither shares Maltese's canary diacritics or clitic-hyphen
morphology; Hungarian needed no morphological joiner, and the
Luxembourgish domain differs from either synthetic-Maltese or
synthetic-Hungarian training data.

Hungarian shows no statistically resolved improvement over stock
Tesseract ($p=0.656$; 95\% CI from -0.00371 to 0.00543).
Luxembourgish improves substantially over the stock configuration
($p<0.0001$; 95\% CI 0.07623--0.09642). The Luxembourgish comparison is
against our stock configuration rather than the published 12.12\% CER
benchmark \cite{agbetimessan2026bnlbenchmark}, which uses a larger,
differently sampled test set and a substantially stronger baseline.

\begin{table}[H]
\caption{Cross-language results.}
\label{tab:crosslang}
\resizebox{\columnwidth}{!}{%
\begin{tabular}{@{}llrll@{}}
\toprule
Language & Stage & CER & $p$ \\
\midrule
Hungarian & stock Tesseract & 0.13543  & \\
Hungarian & five-stream arbitration & 0.13441 & \\
Hungarian & full pipeline & 0.13438 & $0.656$ \\
Luxembourgish & stock Tesseract & 0.25520  & \\
Luxembourgish & five-stream arbitration & 0.17413 & \\
Luxembourgish & full pipeline & 0.16927 & $<0.0001$ \\
\bottomrule
\end{tabular}}
\end{table}

\subsection{Released Maltese OCR Corpus}
\label{sec:corpus}

Paragraph-scale Maltese OCR training data remains scarce; NOMOCRAT
describes 57 annotated pages~\cite{tanti2023nomocrat}, alongside the
competition's own resources. We built and released a public, reusable corpus:
36{,}803
paragraph-level (image, text) pairs, 94 from EUR-Lex legal documents and
36{,}709 from Maltese Wikipedia\footnote{\url{https://huggingface.co/datasets/radmada/maltese-ocr-corpus}},
with per-sample source and licence provenance. Pages are rendered at 150~DPI;
paragraph regions come from
PyMuPDF's\footnote{\url{https://pymupdf.readthedocs.io/}} text-block
segmentation of the
PDF's own layout, not its Unicode text layer, whose ToUnicode maps are
unreliable for Maltese diacritics in these born-digital documents. Each
crop's label is instead aligned against an alternate representation of the
same content, EUR-Lex's Formex XML or Wikipedia's plaintext-extract API,
and kept only when word-level substring coverage against that source
reaches 0.7 (per-pair score provided in the metadata); the two sources are
alternate renderings of the same document, not independent authorities on
its correctness.

The corpus is dominated by encyclopedic prose: 99.7\% of pairs are
derived from Wikipedia, with legal and administrative Maltese therefore
underrepresented. The documents are born-digital rather than scanned,
so the corpus captures real text and layout but not scanner noise or
historical-document degradation. It was created after the competition
and did not contribute to the reported held-out result.

\section{Limitations}
\label{sec:threats}

The DocEng result establishes performance on a held-out Maltese
benchmark under the competition's fixed execution environment.
Generalisation beyond that setting is less directly established. The
training procedure relies heavily on synthetic rendering, while the
released real-text corpus consists primarily of born-digital Wikipedia
material rather than degraded scanned documents. Cross-language
transfer was also mixed: Luxembourgish improved substantially over the
stock configuration, whereas Hungarian did not show a significant
difference. Evaluation on larger independently collected Maltese OCR
datasets, particularly historical and degraded scans, would provide a
stronger test of generalisation.

\section{Conclusion}
\label{sec:conclusion}

LV-ROVER-MLT combines synthetic Tesseract fine-tuning with complementary
recognition streams and Maltese-specific lexicon-gated arbitration.
Each stage improves the development-set result, with the complete
pipeline reaching CER~0.00700 from the organizers' 0.0234 Tesseract
baseline.

On the held-out DocEng~2026 Maltese OCR benchmark, LV-ROVER-MLT placed
first with CER~0.0074. The next-ranked submission scored 0.0161 and
NOMOCRAT scored 0.0163. The result shows that synthetic fine-tuning and
complementary Tesseract configurations can provide effective low-resource
OCR when their hypotheses are combined with language-specific
constraints.

The same approach improved Luxembourgish relative to the stock
configuration, while the Hungarian result was inconclusive. The
released 36,803-pair Maltese OCR corpus provides additional
paragraph-level data for evaluation and future model development.

\section{Availability}
\label{sec:reproducibility}

The inference code\footnote{\url{https://github.com/adamd1985/lv-rover-mlt}}
is Apache~2.0. The weights\footnote{\url{https://huggingface.co/radmada/lv-rover-mlt}}
train on korpus\_malti text (CC~BY-NC-SA~4.0, access-gated); the released
weights are distributed under the same CC~BY-NC-SA~4.0 restriction. The corpus
(Section~\ref{sec:corpus}) is per-sample licensed: EUR-Lex under
Commission Decision 2011/833/EU, Wikipedia text under CC~BY-SA~4.0.

\bibliographystyle{ACM-Reference-Format}
\bibliography{paper}

\clearpage
\appendix

\section{LV-ROVER-MLT Inference Algorithm}
\label{app:algorithm}

Algorithm~\ref{alg:lvrover} specifies LV-ROVER-MLT inference. $P_j$ is
the joined, normalised paragraph from stream $j$ and $W_j$ its
word-tokenised form; $L$ is the Maltese lexicon; $\mathrm{ed}$ is
Levenshtein distance; $\mathrm{alpha}$, $\mathrm{nonascii}$ and
$\mathrm{diac}$ count alphabetic, non-ASCII alphabetic, and Maltese
canary characters (\textit{ċ ġ ħ ż} and capitals) respectively.

\begin{algblock}{LV-ROVER-MLT per-image inference.}
\label{alg:lvrover}
\footnotesize
\noindent\textbf{Input:} image $I$, lexicon $L$\quad\textbf{Output:} paragraph string $s$
\smallskip

\begin{tabbing}
\hspace{1.0em}\=\hspace{1.0em}\=\hspace{1.0em}\=\kill
\textbf{for} $j \in \{1,\ldots,5\}$ \textbf{do} \quad\textit{(recognise, join, normalise)}\\
\>$I_j \gets \textsc{Upscale}(I,2{\times})$ if $j{=}5$ else $I$\\
\>$P_j \gets \textsc{Join}(\textsc{Tesseract}(I_j,\mathrm{config}_j))$, hyphen-repaired, NFC\\
\textbf{end for}\\[2pt]
$P^* \gets P_2$; \textbf{if} $|P^*| < 0.6\max_j|P_j|$ \textbf{then} $P^* \gets \arg\max_j|P_j|$\\[2pt]
\textbf{if} $|P^*|\ge100$: \textbf{for} each $w\notin L$ with $|w|\ge3$,\\
\>rank single-character substitutions by $P(\mathrm{true}\mid\mathrm{tess})$ descending;\\
\>accept the first $c\in L$ with $\mathrm{alpha}(c)\ge\mathrm{alpha}(w)$,\\
\>$\mathrm{nonascii}(c)\ge\mathrm{nonascii}(w)$, $\big||c|{-}|w|\big|\le1$, $\mathrm{ed}(w,c)\le2$\\[2pt]
Candidates $\gets$ the non-anchor streams, identical complete paragraphs\\
\>collapsed; align each distinct candidate's $W_j$ to anchor $W^*$;\\
\>unmatched candidate insertions are discarded (anchor structure kept)\\
\textbf{for} each anchor word $a$ at position $p$ \textbf{do}\\
\>\textbf{for} each aligned candidate $c$, in stream order \textbf{do}\\
\>\>\textbf{if} $\mathrm{strip\_diac}(a){=}\mathrm{strip\_diac}(c)$, $\mathrm{diac}(c)>\mathrm{diac}(a)$,\\
\>\>$c\in L$, $\mathrm{ed}(a,c)\le2$ \textbf{then} propose $c$ \quad\textit{(restoration gate)}\\
\>\>\textbf{else if} $a\notin L$, $c\in L$, $\mathrm{alpha}(c)\ge\mathrm{alpha}(a)\ge3$,\\
\>\>$|c|\ge|a|{-}1$, $\big||a|{-}|c|\big|\le2$, $\mathrm{nonascii}(c)\ge\mathrm{nonascii}(a)$,\\
\>\>$1\le\mathrm{ed}(a,c)\le2$ \textbf{then} propose $c$ \quad\textit{(ordinary swap gate)}\\
\>\textbf{end for}\\
\>$\hat{W}[p] \gets$ the proposal with the most votes, ties broken by the\\
\>\quad fixed candidate iteration order; else $a$ if no proposal\\
\textbf{end for}\\[2pt]
$s \gets \textsc{CurlDoubleQuote}(\textsc{CurlApostrophe}(\textsc{FixLeadMarker}(\textsc{Join}(\hat{W}))))$\\
\textbf{return} $s$
\end{tabbing}
\end{algblock}

\section{Synthetic Data Generation Algorithm}
\label{app:synth-algorithm}

Algorithm~\ref{alg:synth} specifies the synthetic rendering procedure
used to generate training samples.

\begin{algblock}{Synthetic paragraph generation, per sample.}
\label{alg:synth}
\footnotesize
\noindent\textbf{Input:} corpus paragraph text $t$, validated 68-face font
pool $F$ (62 printed, 6 handwriting), augmentation config
\quad\textbf{Output:} image $I$, label $\ell$
\medskip

\begin{tabbing}
\hspace{1.1em}\=\hspace{1.1em}\=\hspace{1.1em}\=\kill
\textit{Stage 1: layout}\\
\>font $\gets$ sample($F$); pt $\gets$ uniform(8, 14);\\
\>width $\gets$ uniform(400px, 1200px)\\
\>lines $\gets$ greedy word-wrap($t$, font, width)\\
\>\textbf{if} lines$[-1]$ ends mid-compound \textbf{then} rewrap before the\\
\>\quad dash, not after \quad\textit{(open-class compounds, not clitics)}\\[3pt]
\textit{Stage 2: hyphenation and dash tagging}\\
\>\textbf{for} each line break, with probability $0.06$ \textbf{do}\\
\>\>split trailing word; tag U+00AD at the label join point\\
\>\>classify the printed hyphen: soft / structural / compound\\
\>\textbf{end for}\\
\>with probability $0.30$, redraw a printed em-dash as an\\
\>\quad image-only en-dash (label keeps U+2014 regardless)\\[3pt]
\textit{Stage 3: render}\\
\>draw lines onto a white canvas at the corrected DPI\\
\>\quad (half-resolution Lanczos rescale, Section~\ref{sec:renderer-dpi})\\
\>justify with probability $0.45$; vary line spacing,\\
\>\quad padding, and leading-bullet markers\\[3pt]
\textit{Stage 4: augment}\\
\>apply, in order: rotation ($\pm 1.5^\circ$), Gaussian blur,\\
\>\quad brightness/contrast jitter, ink bleed, column-edge crop,\\
\>\quad mild elastic distortion, salt-and-pepper, JPEG re-encode\\
\>\quad (quality 65--80 jitter around the 72 calibration target,\\
\>\quad matching real-crop compression)\\
\>\textbf{return} augmented image $I$, label $\ell$ (soft hyphens as\\
\>\quad U+00AD, structural hyphens as literal \texttt{-})
\end{tabbing}
\end{algblock}

Rendered line samples are accumulated into training shards and used to
warm-start Tesseract's stock Maltese model on the merged character set;
model training and checkpoint selection are performed separately from
the per-sample rendering procedure.

\end{document}